\documentclass[final]{article}

\usepackage[nonatbib]{neurips_2024}

\usepackage[utf8]{inputenc} 
\usepackage[T1]{fontenc}    
\usepackage{hyperref}       
\usepackage{url}            
\usepackage{booktabs}       
\usepackage{amsfonts}       
\usepackage{nicefrac}       
\usepackage{microtype}      
\usepackage{xcolor}         

\usepackage{listings}
\usepackage{amsmath}
\usepackage{graphicx}
\usepackage{subfig}
\usepackage[
    backend=biber,
    natbib=true,
]{biblatex}
\title{Minimal Time Series Transformer}

\author{%
  Joni-Kristian K{\"a}m{\"a}r{\"a}inen\thanks{See \url{https://webpages.tuni.fi/vision/public_pages/JoniKamarainen/}}\\
  Department of Computing Sciences\\
  Tampere University
}

\begin{document}

\maketitle

\begin{abstract}
Transformer is the state-of-the-art model for many natural language processing, computer vision, and audio analysis problems. Transformer effectively combines information from the past input and output samples in autoregressive manner so that each sample becomes aware of all inputs and outputs. In sequence-to-sequence (Seq2Seq) modeling, the transformer processed samples become effective in predicting the next output. Time series forecasting is a Seq2Seq problem. The original architecture is defined for discrete input and output sequence tokens, but to adopt it for time series, the model must be adapted for continuous data. This work introduces minimal adaptations to make the original transformer architecture suitable for continuous value time series data.
\end{abstract}

\begin{center}
\texttt{Jupyter notebook:} \url{https://github.com/kamarain/minimal_time_series_transformer}
\end{center}

\section{Introduction}
In their seminal work, Vaswani~\textit{et~al.}~\cite{transformer} introduced
the transformer architecture for machine translation. The proposed model
was trained supervised manner with training data consisting input $\mathbf{X}$ and output sequences $\mathbf{Y}$. The sequences were corresponding sentences in two languages. During inference, the model is executed
in generative manner
\begin{displaymath}
  \begin{split}
    \hat{y}_0 &= \hbox{Transformer}(\mathbf{X},SOS),\\
    \hat{y}_1 &= \hbox{Transformer}(\mathbf{X},SOS,\hat{y}_0),\\ 
    \hat{y}_2 &= \hbox{Transformer}(\mathbf{X},SOS,\hat{y}_0,\hat{y}_1),\\
    \ldots \\
    \hat{y}_N &= \hbox{Transformer}(\mathbf{X},SOS,\hat{y}_0,\hat{y}_1,\ldots,\hat{y}_{N-1}) = EOS    
  \end{split}
  \label{eq:seq2seq}
\end{displaymath}
to produce an output sequence $\mathbf{Y} = \left[ SOS, \hat{y}_0, \ldots, \hat{y}_N \right]$
(translation) corresponding to the input sequence $\mathbf{X}$, and in which SOS and EOS are
special symbols (tokens) 'start-of-sequence' and 'end-of-sequence'. In follow up works, the transformer model has been extended to other modalities and tasks.
\textit{Transformer} is undoubtly the state-of-the-art model for
sequential data.

Time series are used in statistics, signal processing, econometrics, control engineering, and largely in any domain of applied science and engineering which involves temporal measurements.
\textit{Time series forecasting} is the use of a time series model to predict future values based on previously observed values. The forecasting problem can be cast into the sequence-to-sequence (Seq2Seq) model and generative inference defined in (\ref{eq:seq2seq}). Therefore it is intriguing to adopt transformers for time series forecasting.

Several adaptations have been proposed for time series forecasting.
For
example, LogSparse Transformer~\cite{Li-2019-neurips}, MTS Transformer~\cite{Zerveas-2021-kdd}, Informer~\cite{Informer}, and ContiFormer~\cite{ContiFormer}. These works address various challenges of time series forecasting such as varying sampling rate and long-term dependencies. However, it is not clear what are the minimal changes to adapt the 'vanilla' transformer~\cite{transformer} for sequences of continuous data. Some adaptation is needed since the vanilla transformer is designed for discrete inputs and outputs, tokens, and does not work with continuous data. The rather complicated adaptations in the previous works raise questions since they have not been compared to any 'time series transformer baseline'. Moreover, since Zeng~\textit{et~al.}~\cite{Zeng-2023-aaai} experimentally demonstrated that an ``embarassingly simple'' single layer linear regressor outperforms state-of-the-art time series transformers in multiple benchmarks, it is justified to define a simple time series transformer baseline - \textit{a minimal time series transformer}.

This work investigates minimal adataptions of the vanilla transformer~\cite{transformer} for time series forecasting. To keep things simple and understandable, all experiments are performed on Sinusoid sequences. All code for the time series transformer models and experiments are made publicly available.


\section{Methods}

\subsection{Vanilla Transformer - Seq2SeqTransformer}
Our reference implementation is the experimentally verified \texttt{Seq2SeqTransformer} class
from~\cite{Kamarainen-2025-transintro}. Their class uses the popular PyTorch
\texttt{torch.nn.Transformer} class as the main building block. This makes the \texttt{Seq2SeqTransformer}
\texttt{forward()} function easy to follow:
\begin{lstlisting}
.forward()
        # Note: src & tgt default size is (seq_length, batch_num, feat_dim)

        # Token embedding
        src = self.embedding(src) * math.sqrt(self.d_model)
        tgt = self.embedding(tgt) * math.sqrt(self.d_model)

        # Positional encoding - _must_ be seq len x batch num x feat dim
        # Inference often misses the batch num
        if src.dim() == 2: # seq len x feat dim
            src = torch.unsqueeze(src,1) 
        src = self.positional_encoder(src)
        if tgt.dim() == 2: # seq len x feat dim
            tgt = torch.unsqueeze(tgt,1) 
        tgt = self.positional_encoder(tgt)

        # Transformer output
        out = self.transformer(src, tgt, tgt_mask=tgt_mask,
                               src_key_padding_mask = src_key_padding_mask,
                               tgt_key_padding_mask=tgt_key_padding_mask,
                               memory_key_padding_mask=src_key_padding_mask)
        out = self.unembedding(out)
        
        return out

\end{lstlisting}

The main processing steps are 1) the source and target sequence embedding using the PyTorch \texttt{torch.nn.Embedding} class that maps token ids (integer values) to one-hot encoded vectors and further to an embedding vector,
2) positional encoding implemented using an encoding class that follows the original work~\cite{transformer},
and 3) the linear 'un-embedding' layer that transforms the transformer outputs to class probabilities. Note that the PyTorch implementation of cross-entropy loss adds the softmax-nonlinearity, and thus it is not explicitly in the forward function. The PyTorch transformer class takes care of masking the future outputs and padding keys, and mask construction is available in the reference code. A general, and often confusing, requirement is that the source and target variables must be three dimensional:
\textit{sequence length} $\times$
\textit{number of batches} $\times$
\textit{feature vector dimension}.
Inside the class there are some cumbersome steps to maintain the correct shapes, and these steps are often the causes of programming bugs.

\subsection{Minimal time series transformer - MiTS-Transformer}
\label{sec:minimal}
The minimal adaptation to the discrete token \texttt{Seq2SeqTransformer} is to change the 'integer-to-vector' embedding layer (\texttt{torch.nn.Embedding}) to a layer that converts continuous value vectors to vectors of the model dimension - 'vector-to-vector'. A neural network trick to do this is to replace the embedding layer with a linear layer. This can be done by a small change in the original code. The original embedding
\begin{lstlisting}
.init()
  self.embedding = nn.Embedding(num_tokens, d_model)

  self.unembedding = nn.Linear(d_model, num_tokens)
\end{lstlisting}
is replaced by
\begin{lstlisting}
.init()
  self.embedding = nn.Linear(d_input, d_model)

  self.unembedding = nn.Linear(d_model, d_input)
\end{lstlisting}
The embedding in the continuous case maps the \texttt{d\_input}-dimensional sample to \texttt{d\_model}-dimensional
model vector. In the un-embedding step the conversion is made backwards.

\subsection{Positional encoding expansion - PoTS-Transformer}
\label{sec:excodat}
In time series forecasting, there are three potential challenges for transformers,
\textit{i)} sequences can be long, from thousands to tens of thousands samples,
\textit{ii)} temporally close samples can be highly correlated, and
\textit{iii)} amount of training data can be limited.
The straightforward solutions to the challenges are conflicting. Long sequences require high dimensional
models so that there is 'space' for position information. The small amount of data makes it difficult to train large models due to overfitting. Strong correlation between the adjacent samples makes it possible to use small models since each new sample provides only negligible amount of new information. To meet the conflicting requirements, special tricks are needed.

In the existing literature, particularly the long sequences have received attention. For example, Li~\textit{et~al.}~\cite{Li-2019-neurips} propose logarithmically sparse sampling of sequences so that pose encoding becomes logarithmic reducing the need of high dimensional models. Wav2vec 2.0 is a state-of-the-art audio backbone network for audio feature extration~\cite{wac2vec2}. It compresses long audio sequences to more compact representation - 'audio tokens' - by first applying convolution filters and then using a nearest-neighbor product quantizer~\cite{Jegou-2011-pami}.

What is the simplest solution for the challenges? In \textit{PoTS-Transformer} the model size is kept low to avoid overfitting. At the same time, the positional encoding is done in a higher dimensional space to keep it effective for long sequences. This is implemented by wrapping the positional encoder between two linear layers that first perform \textit{positional encoding expansion} and after the encoding step \textit{inverse positional expansion}:
\begin{lstlisting}
  .forward()
        ...
        src = self.pos_expansion(src)
        src = self.positional_encoder(src)
        src = self.pos_invexpansion(src)
        ...
        tgt = self.pos_expansion(tgt)
        tgt = self.positional_encoder(tgt)
        tgt = self.pos_invexpansion(tgt)
\end{lstlisting}
The expansion and its inverse are defined as
\begin{lstlisting}
  .init()
        ...
        # Positional encoding expansion
        self.pos_expansion = nn.Linear(d_model, pos_expansion_dim)
        self.pos_invexpansion = nn.Linear(pos_expansion_dim, d_model)
        

        # Positional encoding
        self.positional_encoder = PositionalEncoding(d_model=pos_expansion_dim,
                                                     dropout=dropout_p)
        ...
\end{lstlisting}

Performing the transformer model computations with 8-dimensional vectors and the pose encoding with 128-dimensional vectors increases the number of learnable parameters in PoTS-Transformer from 1,433 to 3,473 ($2.4\times$). For comparison, increasing the model size of MiTS-Transformer from 8- to 128-dimensional spaces increases the number of parameters from 1,289 to 204,689 ($158\times$). The difference is nearly two orders of magnitude.

\subsection{Training details}
All transformer models were trained with the standard PyTorch Adam optimizer (\texttt{torch.optim.Adam}) using the initial learning rate 0.023. A multistep scheduler was used to decay the learning rate of each parameter group by gamma once the number of epoch reaches one of the milestones. The gamma was fixed to 0.1 and the milestones were manually optimized for each case separately, and the goal losses documented to facilitate replication of all experiments. However, similar results can be achieved with the fixed learning rate 0.023, no scheduling, and training for 2000 epochs. The training of the largest model in our experiments took only a few minutes on a standard laptop without a GPU.

\section{Data}
\label{sec:data}

\begin{figure}[h]
  \centering
  \subfloat[f=0/31]{\includegraphics[width=0.25\linewidth]{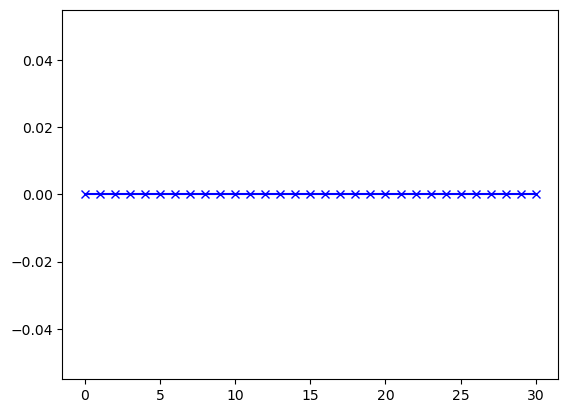}}
  \subfloat[f=1/31]{\includegraphics[width=0.25\linewidth]{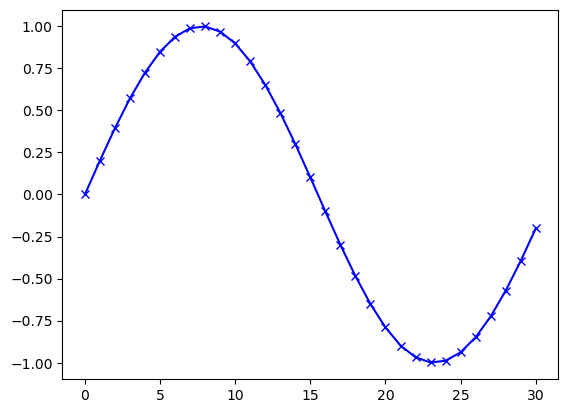}}
  \subfloat[f=2/31]{\includegraphics[width=0.25\linewidth]{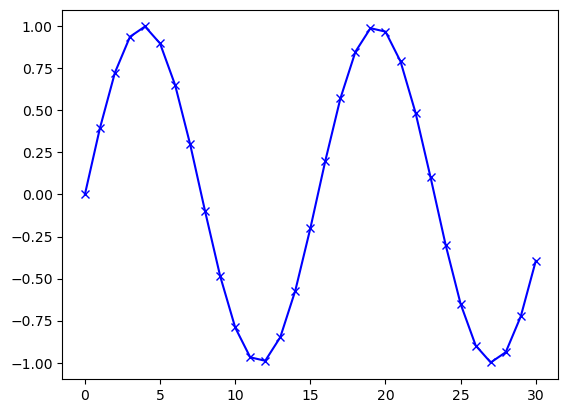}}
  \subfloat[f=3/31]{\includegraphics[width=0.25\linewidth]{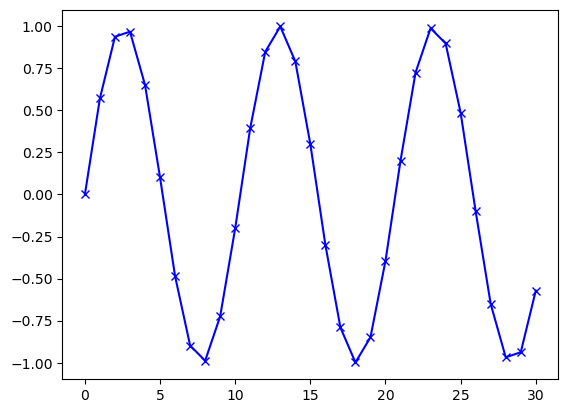}}
  \caption{Example sinusoids used in the experiments.\label{fig:sin_examples}}
\end{figure}

The Seq2Seq data for the experiments were generated by sampling the sine function
\[
y(t) = \sin{2\pi ft}
\]
$f$ is the since frequency and $t$ defines the time instant. For discrete signals, it is convenient to define the
frequency as discrete frequency of waveforms per number of samples:
\[
 f = \frac{w}{L}
\]
where $w \in \mathbb{R}$ is the number of waves and $L$ the number of samples. For example, $f=2/31$ means
2 waveforms per 31 samples. Sinusoids with varying frequencies are shown in Figure~\ref{fig:sin_examples}. Figure~\ref{fig:sin_examples}(d) shows sampling errors that become more dominant as the frequency increases (note that the top of each wave looks different).

\begin{figure}[h]
  \centering
  \subfloat[f=0/31]{\includegraphics[width=0.25\linewidth]{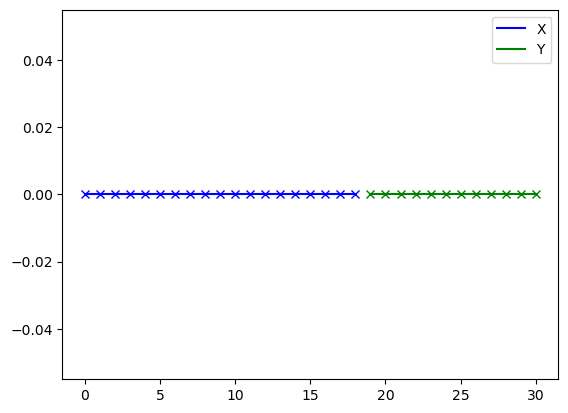}}
  \subfloat[f=1/31]{\includegraphics[width=0.25\linewidth]{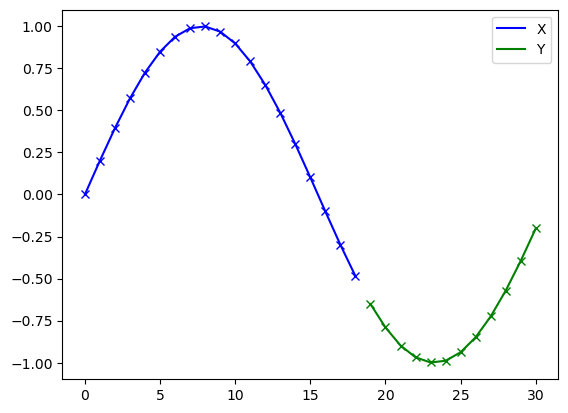}}
  \subfloat[f=2/31]{\includegraphics[width=0.25\linewidth]{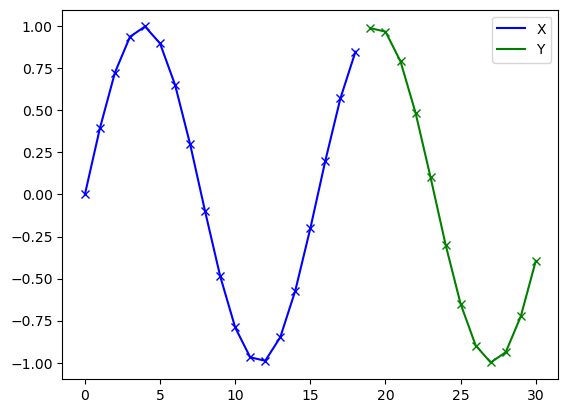}}
  \subfloat[f=3/31]{\includegraphics[width=0.25\linewidth]{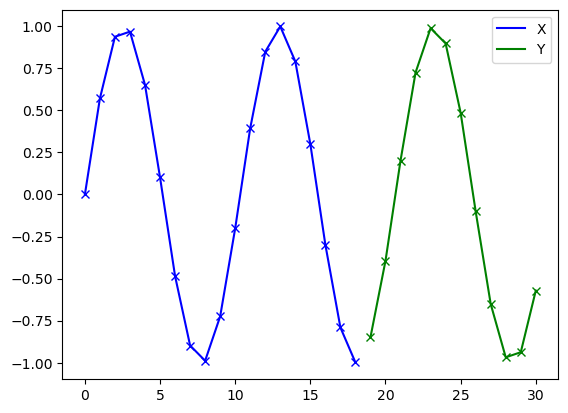}}
  \caption{Sinusoids (Type 2 sequences) of 31 samples divided into source $\mathbf{X}$ (samples 0-18 in blue) and target $\mathbf{Y}$  parts (samples 19-30 in green).\label{fig:type2_data}}
\end{figure}

\paragraph{Data types.} Three types of data were generated:
\begin{enumerate}
\item Type 1: Single sequence (sinusoid with $f=1/L$)
\item Type 2: A fixed number of sequences (e.g., $f=0/L,~1/L,~2/L,~3/L$)
\item Type 3: Arbitrary number of sequences ($f \in U(0,f_{max})$)
\end{enumerate}
$U()$ is the uniform random distribution. Type 1 is used to sanity check the models. Type 2 is an easy case where there are a small number of different sequences. Type 3 is the most difficult case as the difference between two sinusoids of the nearly same frequency can be subtle.
In the experiments the signal length is set to 31 which is divided to 19 input (source) samples $\mathbf{X}$ and 12 output (target) samples $\mathbf{Y}$ (Figure~\ref{fig:type2_data}).

\section{Experiments}

\subsection{Code sanity check with a single fixed sequence (Type 1 sequences)}
The sanity checks were done with a single sinusoid of the lenght $L=31$ and frequency $f=1/31$. The signal was repeated 100 times in the training and test sets to mimic standard training.

\begin{figure}[h]
  \centering
  \subfloat[100 epochs, loss 0.016, err 4.13]{\includegraphics[width=0.45\linewidth]{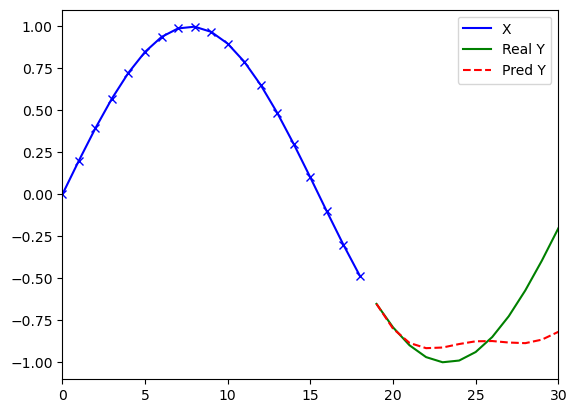}}
  \subfloat[200 epochs, loss 0.003, err 0.23]{\includegraphics[width=0.45\linewidth]{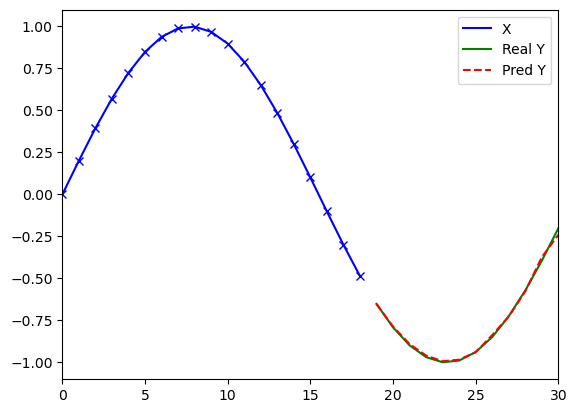}}
  \caption{Single sequence sanity check (Type 1 data) of the \texttt{MiTS-Transformer} implementation (see Jupyter notebook).\label{fig:sanity_MiCoDaT}}
\end{figure}

\paragraph{MiTS-Transformer.}
The model converged after 200 epochs with the fixed learning rate 0.023 and obtained the final MSE error 0.23. See Figure~\ref{fig:sanity_MiCoDaT} for examples.

\subsection{Multiple sequences (Type 2)}

\begin{figure}[h]
  \subfloat[100 epochs, loss 0.056 err 7.30]{\includegraphics[width=0.45\linewidth]{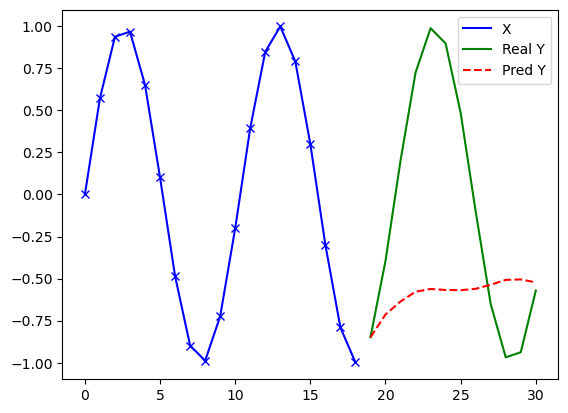}}
  \subfloat[600 epochs, loss 0.012 err 0.61]{\includegraphics[width=0.45\linewidth]{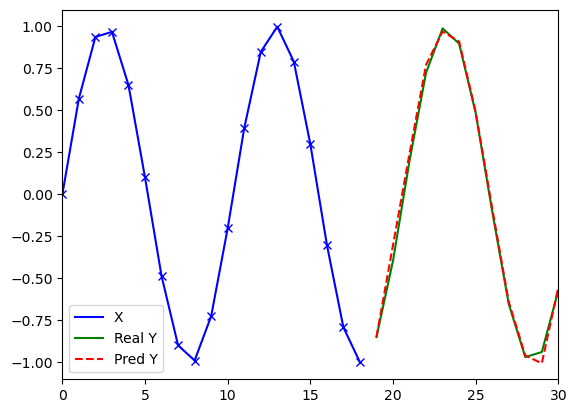}}
  \caption{Four sequence (Type 2) results of \texttt{MiTS-Transformer} (model parameters: \texttt{d\_model=8, dim\_feedforward=8}, the total of 1,289 learnable parameters). The signal frequencies were 0/31, 1/31, 2/31, and 3/31.\label{fig:type2_MiCoDaT}}
\end{figure}

\paragraph{MiTS-Transformer.} The results and illustrations in Figure~\ref{fig:type2_MiCoDaT} demonstrate good learning performance. There is virtually no difference between the single and multiple (4) sequence accuracy.

\subsection{Arbitrary sequences (Type 3)}

\begin{figure}[h]
  \subfloat[R1: loss 0.018 err 4.32]{\includegraphics[width=0.33\linewidth]{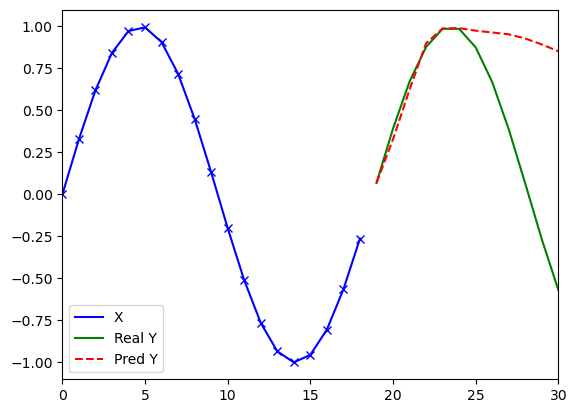}}
  \subfloat[R2: loss 0.019 err 3.28]{\includegraphics[width=0.33\linewidth]{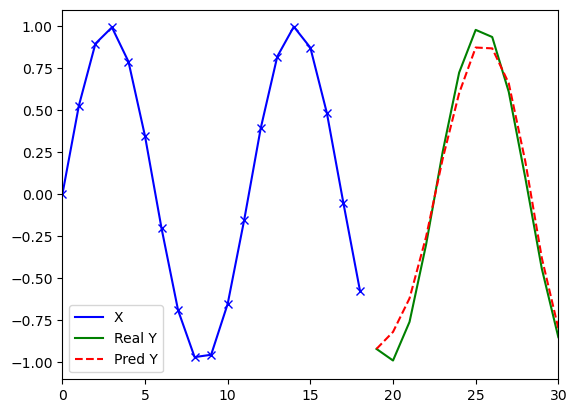}}
  \subfloat[R3: loss 0.021 err 3.37]{\includegraphics[width=0.33\linewidth]{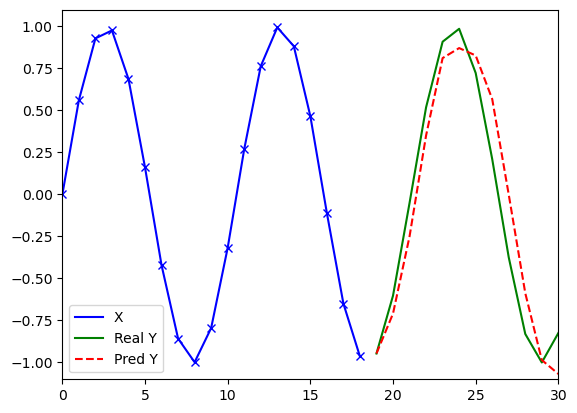}}
  \caption{Arbitrary sequences (Type 3) results for \texttt{MiTS-Transformer} (\texttt{d\_model=8, dim\_feedforward=8}, 1289 params). Data consists of arbitrary sequences of Sinusoids with the frequency in (0/31, 3/31).\label{fig:type3_MiCoDaT8}}
\end{figure}

\paragraph{MiTS-Transformer.} The results and sample illustrations from three different runs with the same training and test data are in Figure~\ref{fig:type3_MiCoDaT8}. The results verify that the minimal time series transformer works for data with arbitrarily small changes in the sequences. The results verify certain \textit{interpolation capability} of the model since many test sequences do not appear in training data. The results for the arbitrary sequences are almost order of magnitude worse than for the fixed Type 1 and 2 sequences.

\begin{figure}[h]
  \subfloat[R1: loss 0.008 err 3.16]{\includegraphics[width=0.33\linewidth]{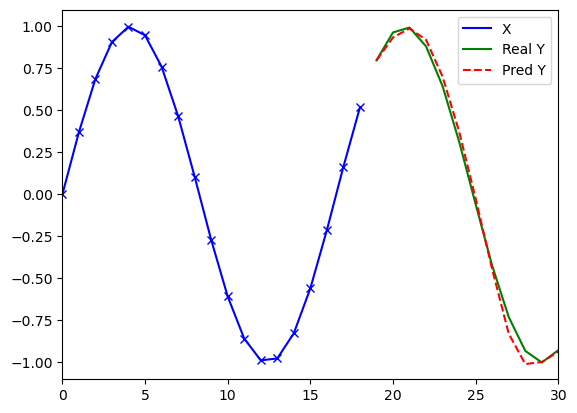}}
  \subfloat[R2: loss 0.010 err 2.66]{\includegraphics[width=0.33\linewidth]{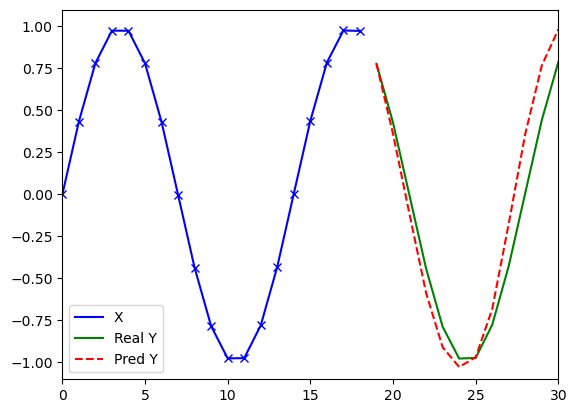}}
  \subfloat[R3: loss 0.010 err 2.29]{\includegraphics[width=0.33\linewidth]{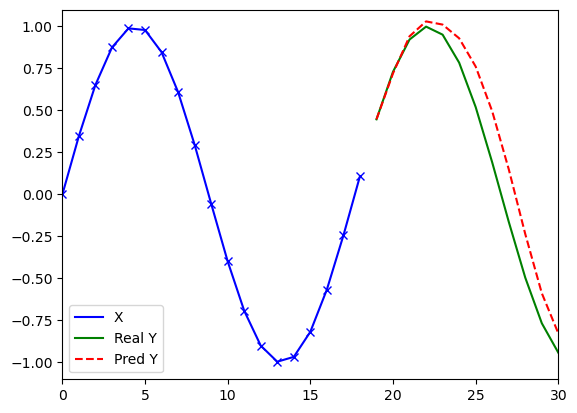}}
  \caption{Arbitrary sequences (Type 3) results for \texttt{MiTS-Transformer} (\texttt{d\_model=16, dim\_feedforward=8}, 4097 params).\label{fig:type3_MiCoDaT16}}
\end{figure}

\begin{figure}[h]
  \subfloat[R1: loss 0.004 err 1.46]{\includegraphics[width=0.33\linewidth]{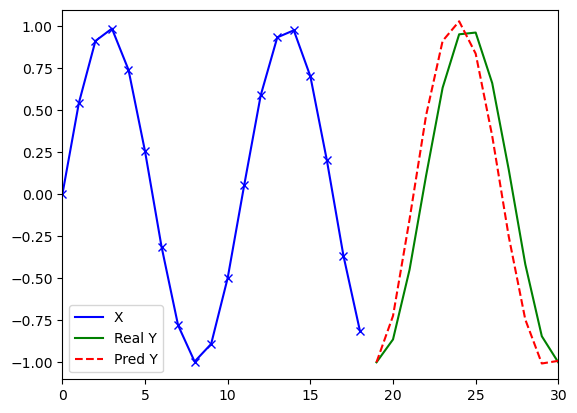}}
  \subfloat[R2: loss 0.011 err 3.02]{\includegraphics[width=0.33\linewidth]{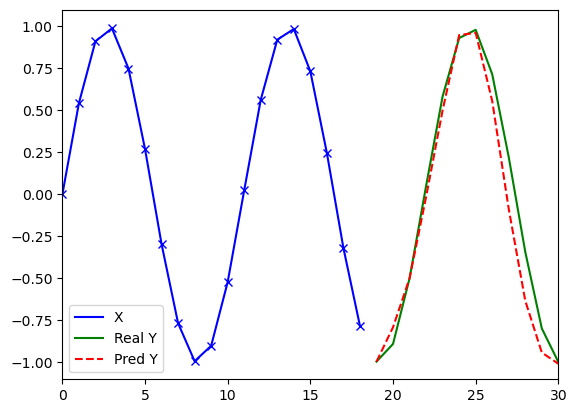}}
  \subfloat[R3: loss 0.006 err 2.76]{\includegraphics[width=0.33\linewidth]{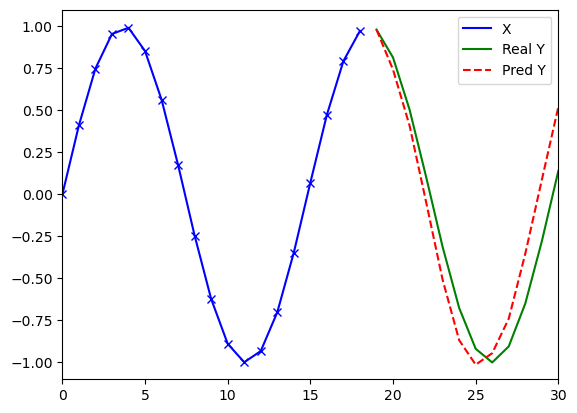}}
  \caption{Arbitrary sequences (Type 3) results for \texttt{MiTS-Transformer} (\texttt{d\_model=32, dim\_feedforward=8}, 14321 params).\label{fig:type3_MiCoDaT32}}
\end{figure}

The learning capacity of MiTS-Transfomer can be increased by increasing the model dimension (\texttt{d\_model}). In Figures~\ref{fig:type3_MiCoDaT16} and \ref{fig:type3_MiCoDaT32} are results for the models with the model dimension set to 16 and 32. The size 16 model is systematically better than the size 8 model, but the size 32 model starts to occasionally overfit to the training data. The total number of learnable parameters in the 8, 16, and 32 models are 1289 to 4097 and 14321, respectively.

\begin{figure}[h]
  \subfloat[R1: loss 0.004 err 1.31]{\includegraphics[width=0.33\linewidth]{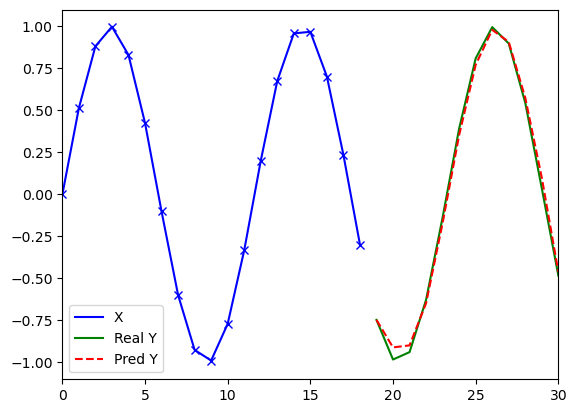}}
  \subfloat[R2: loss 0.006 err 1.49]{\includegraphics[width=0.33\linewidth]{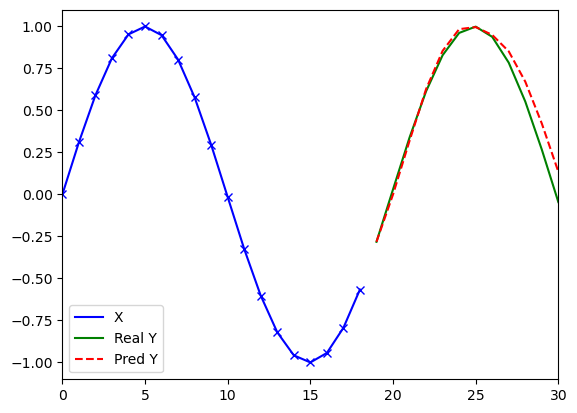}}
  \subfloat[R3: loss 0.005 err 1.45]{\includegraphics[width=0.33\linewidth]{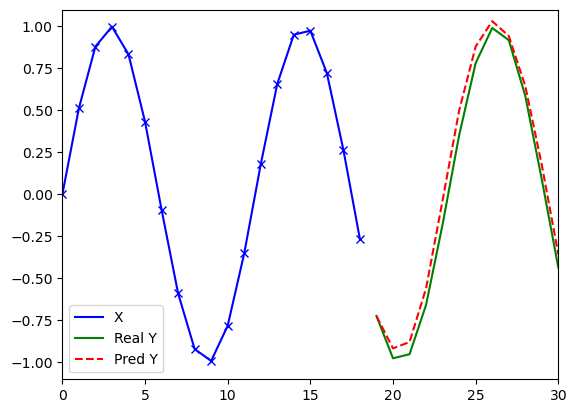}}
  \caption{Arbitrary sequences (Type 3) results for the \texttt{PoTS-Transformer} (\texttt{d\_model=8, dim\_feedforward=8, pos\_expansion\_dim=64}, 2,385 params).\label{fig:type3_PoTS32}}
\end{figure}

\paragraph{PoTS-Transformer.} The experiments were run using the positional expansion dimension 64 and 2,385 learnable parameters. The results in Figure~\ref{fig:type3_PoTS32} demonstrate systematically better performance than the MiTS-Transformer.

\section{Conclusions}
The purpose of this work and the accompanying Jupyter notebook was to study how the ``Attention is All You Need'' (vanilla) transformer for discrete tokens can be adapted for time series (continuous) data. The minimal adaptation is to change the token embedding layer to a linear layer. This was implemented in our minimal time series transformer (MiTS-Transformer) which learned sinusoids very well. The learning capacity of MiTS-Transformer can be adjusted by changing the model dimension. This however quickly leads to model size explosion in the terms of learnable model parameters and the model easily overfits. For this problem a simple model called positional encoding expansion time series transformer (PoTS-Transformer) was proposed. It combines positional encoding of long sequences in the expanded space and a low dimensional model avoiding overfitting. The results in our experiments were convincing and pave way for similar minimal and simple tricks for transformer based time series forecasting.

\begin{ack}
This work would not be possible without my tenure at Tampere University. Tenure allows me to allocate time to projects that interest me. To thank my institute and the Finnish government, I share my results and findings with everyone interested in machine learning and programming.

If you find my writings or code helpful, please cite this work. The work is available as an ArXiv article. Mention the author name(s), title, and the ArXiv URL in your project report, Web page, thesis, etc.

Most universities worldwide are ideal places for curious and playful people like me. I hope the corporative style management will not destroy them as similar people to me have been developing them for the past thousand years.

Keep on Rockin' in the Free World!
\end{ack}

\printbibliography

\end{document}